\documentclass{article}

\usepackage[preprint]{neurips_2026}
\usepackage{multicol}

\usepackage[utf8]{inputenc} 
\usepackage[T1]{fontenc}    
\usepackage[hidelinks]{hyperref}       
\usepackage{url}        
\usepackage{booktabs} 
\usepackage{amsfonts}
\usepackage{nicefrac}       
\usepackage{microtype}     
\usepackage{xcolor}      

\usepackage{amsmath}
\usepackage{amssymb}
\usepackage{mathtools}
\usepackage{amsthm}

\usepackage{algorithm}
\usepackage{algorithmic} 
\usepackage{amsmath}
\usepackage{float} 

\usepackage{microtype}
\usepackage{graphicx}
\usepackage{subcaption}
\usepackage{booktabs}

\usepackage[capitalize,noabbrev]{cleveref}

\title{Bi-semantic Chemical Embedder for Joint Representation Learning of SMILES and Natural Language}

\usepackage{amsmath}

\author{%
  David Ming Segura\textsuperscript{1, 2, $\ast$, $\dagger$} \quad
  Jeremy Goumaz\textsuperscript{1, $\ast$} \quad
  Joshua W. Sin\textsuperscript{1, 3} \\
  \bfseries Bojana Rankovi\'{c}\textsuperscript{1, 2} \quad
  \bfseries Philippe Schwaller\textsuperscript{1, 2, $\dagger$} \\
  \\
  \textsuperscript{1}Laboratory of Artificial Chemical Intelligence (LIAC), EPFL, Lausanne, Switzerland \\
  \textsuperscript{2}National Centre of Competence in Research (NCCR) Catalysis, EPFL, Lausanne, Switzerland \\
  \textsuperscript{3}Process Chemistry \& Catalysis, Synthetic Molecules Technical Development, \\
  F. Hoffmann-La Roche AG, Basel, Switzerland \\
  \\
  \textsuperscript{$\ast$}Equal contribution \quad \textsuperscript{$\dagger$}Corresponding author \\
  \texttt{\{david.segura, philippe.schwaller\}@epfl.ch}
}

\begin{document}

\maketitle

\begin{abstract}
  Transformer models have revolutionized natural language processing (NLP), and text-based molecular representations like SMILES have successfully extended these architectures to chemistry. However, domain-adaptive pre-training often causes models to overfit to chemical syntax, catastrophically forgetting their foundational semantic capabilities. To address this challenge, we introduce CheMatE, a chemistry-oriented embedding model that jointly captures molecular structure and domain-specific natural language within the same representation space. Built on a ModernBERT backbone, CheMatE learns bi-semantic representations through a two-stage training procedure: continued masked language modeling (MLM) followed by a Matryoshka contrastive learning stage via Multiple Negative Ranking Loss (MNRL). First, we train the model using MLM on a novel, large-scale corpus of SMILES-annotated, long-context scientific documents that were constructed and curated from FineWeb and ChemPile (comprising 10.4B and 11.5B tokens, respectively). Subsequently, the model undergoes contrastive learning using a synthetic dataset of SMILES-text pairs algorithmically derived from our original training corpus. This design exposes the model to SMILES-enriched scientific literature, enabling bi-semantic understanding. We evaluate CheMatE across a range of downstream tasks covering molecular property prediction and scientific language understanding. Our results demonstrate that coupling our custom-curated datasets with this sequential training strategy yields robust, highly transferable representations. By effectively unifying structural and contextual signals within a single text-based framework, CheMatE achieves competitive performance across both specialized chemistry models and general-purpose language model baselines.
\end{abstract}

\begin{multicols}{2}
\section{Introduction}

  The rise of machine learning (ML) techniques has led to widespread applications in chemistry problems, supporting chemists in reaction planning, property prediction, optimization, and more generally, in similarity search applications~\cite{coley_machine_2018, bian_generative_2021, jablonka_leveraging_2024, sin_optimization_2025, rankovic_large_2025}. Seminal contributions in the field range from the earliest neural networks for supervised molecular property prediction using graph representations to transformer-based models with molecular text representations for diverse applications. Central to the success of these computational approaches is the choice of underlying data representation. Molecules can be represented through various modalities, from 2D graphs~\cite{montanari_modeling_2020}, 3D conformers~\cite{xu_learning_2021}, structural fingerprints~\cite{rogers_extended-connectivity_2010}, specific natural language descriptions~\cite{edwards_text2mol_2021}, or even through Simplified Molecular Input Line Entry System (SMILES)~\cite{weininger_smiles_1988}. The linearization of molecular structures into SMILES strings unlocked a new spectrum of possibilities and led to the birth of powerful sequence-based modeling techniques~\cite{liu_retrosynthetic_2017, gomez-bombarelli_automatic_2018, schwaller_found_2018, schwaller_molecular_2019}, leveraging the inherent capabilities of transformer architectures with textual data~\cite{vaswani_attention_2017, schwaller_found_2018, devlin_bert_2019}. Additionally, self-supervised transformers have already established a robust paradigm for learning SMILES representations~\cite{wang_smiles-bert_2019, fabian_molecular_2020, irwin_chemformer_2022}. Notably, the RXNFP model successfully maps reaction SMILES into continuous embeddings to capture complex chemical transformations~\cite{schwaller_mapping_2020}. Similarly, the ChemBERTa family of models has set foundational baselines by pre-training on extensive corpora of molecular strings~\cite{chithrananda_chemberta_2020, ahmad_chemberta-2_2022}.
  
  However, while these models serve as strong baselines for SMILES-centered representation tasks, they suffer from two critical limitations. First, models that undergo intensive domain-specific pre-training on these syntactic molecular representations often suffer from domain over-specialization. While they excel at specific featurisation tasks such as property prediction, they struggle to generalize across diverse semantic applications~\cite{chithrananda_chemberta_2020, gururangan_dont_2020, li_revisiting_2024, mukhoti_fine-tuning_2024}. Models pre-trained from scratch on SMILES inherently lack the capability to represent general scientific natural language, while the ones adapted from general language models see this capacity degrade through domain-specific continuous pre-training~\cite{christofidellis_unifying_2023, tan_chemmllm_2025}. Furthermore, as molecular SMILES strings are inherently concise representations of isolated entities, embedding models dedicated to molecular entities are historically optimized for short inputs~\cite{honda_smiles_2019, wang_smiles-bert_2019, chithrananda_chemberta_2020, fabian_molecular_2020, irwin_chemformer_2022}. Consequently, such models are architecturally constrained by a shorter token context limit. 

  We address these limitations through a dual strategy: combining bi-semantic text learning with the integration of long-context documents. By simultaneously mapping SMILES and textual scientific descriptions into a shared semantic space, we avoid our model becoming confined to domain-specific syntax. At the same time, by providing an enlarged context window that handles entire scientific documents rather than isolated strings, our model learns to ground structural chemical tokens within their broader, long-form scientific context. While recent bi-semantic efforts have explored this space for generative applications~\cite{christofidellis_unifying_2023, tan_chemmllm_2025}, the development of encoder-centric models dedicated to bi-semantic representation learning remains largely unexplored.
  
\begin{figure*}[ht]
  \vskip 0.2in
  \begin{center}
    \centerline{\includegraphics[width=\textwidth]{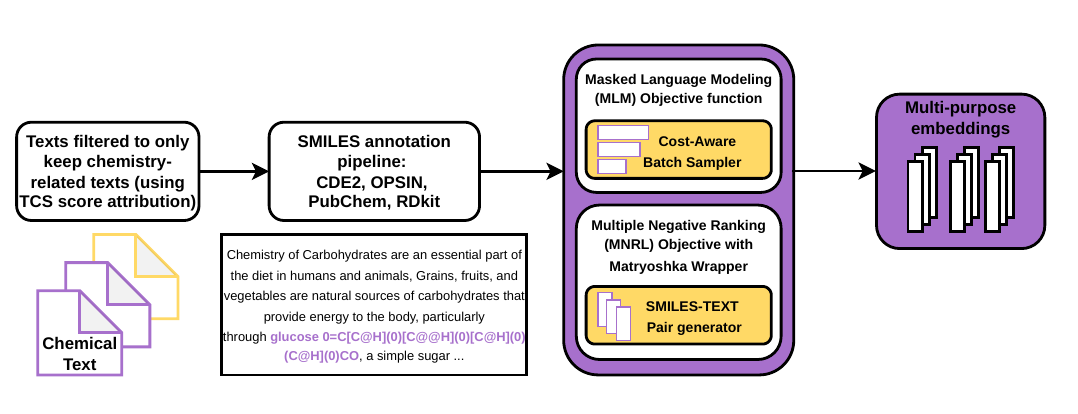}}
    \caption{
      Overview of the multi-task training pipeline for learning bi-semantic chemistry embeddings. Chemical texts are first filtered for domain relevance using TCS score attribution, then processed through a SMILES annotation pipeline (CDE2, OPSIN, PubChem, RDKit) to produce SMILES-injected text corpora. The model is trained sequentially using two objective functions: a Masked Language Modeling (MLM) objective with a Cost-Aware Batch Sampler to improve training efficiency, and a Multiple Negative Ranking (MNRL) objective with a Matryoshka Wrapper with SMILES-TEXT pairs generated from our annotated text. This combined training strategy yields general-purpose embeddings suitable for a range of downstream chemistry tasks. Full training hyperparameters and architectural details are detailed in Section~\ref{sec:implementation_details_mlm} and \ref{sec:implementation_details_cl}.
      }
    \label{fig:pipeline}
  \end{center}
\end{figure*}

  We introduce CheMatE (Chemical Embedder with Matryoshka Embedding), a long-context bi-semantic chemistry encoder built upon a modern transformer architecture~\cite{warner_smarter_2024}. Our model supports sequences up to 8,192 tokens and was extensively trained on an in-house annotated scientific database. Our training strategy relies on a novel, hybrid dataset specifically designed to intertwine molecular structures with rich textual context. We curated more than \textbf{14M long-context documents} (spanning up to 8,192 tokens) primarily consisting of scientific articles and educational materials from FineWeb and ChemPile~\cite{penedo_fineweb_2024, mirza_chempile_2025} (See Figure~\ref{fig:data_distribution}). These texts were then annotated via an automated SMILES-injection pipeline, capable of detecting chemical entities present within the text and inserting a chemical SMILES translation at the next position~\cite{lowe_chemical_2011, mavracic_chemdataextractor_2021, kim_pubchem_2025, landrum2025rdkit}. By tightly weaving structural notation into scientific text, we yield a highly contextualized training corpus directly optimized for bi-semantic representation learning. Ultimately, CheMatE is designed to serve as a versatile foundation for both molecule and natural language embedding applications.
  Across our evaluation pipeline for predictive tasks, CheMatE is the only model that ranks among the best-performing group across both modalities simultaneously, consistently generating high-quality embeddings for both SMILES-based molecular tasks and scientific NLP benchmarks, whereas baseline models typically excel in only one domain. We summarize our core methodological and architectural contributions as follows (Figure~\ref{fig:pipeline}):

  \textbf{(i) Large-scale SMILES injection pipeline.}
  We describe a multi-stage automated pipeline that combines chemical named-entity recognition (NER), structure resolution from public databases, and canonicalization using RDKit~\cite{landrum2025rdkit}, applied across $\sim$14.4~million deduplicated documents drawn from various complementary scientific sources (see Section \ref{sec:data_annotation}). Our pipeline is applied at a large scale, producing a pre-training corpus of 21.9~billion tokens.

  \textbf{(ii) Cost-budget distributed batch sampler.}
  We develop a new batch sampler (\emph{BalancedTokenBatchSampler}) to address the severe length variability across our heterogeneous dataset collection. Rather than enforcing a fixed number of samples per batch, which is not efficient in our case at large-scale with a multi-GPU setup, or ensuring batch homogeneity by sorting the texts by token counts, which would bias the training and reduce batch variety, our sampler enforces three simultaneous constraints for the batch construction: a raw-token budget, a padded-token memory cap, and a quadratic cost ceiling. These constraints are combined with a Distributed Data Parallel (DDP) aware group-balancing policy that guarantees that all GPU ranks process comparable workloads per synchronization step (see Algorithm~\ref{alg:smart_batching}). This reduces mean padding overhead by up to 50\% relative to naive batching and eliminates out-of-memory errors on extremely uneven sequences within a single batch, enabling efficient training at an 8,192-token context length with up to a 30\% increase in GPU Utilization across multi-node GH200 hardware.
  
  \textbf{(iii) Bi-semantic sequential training.}
  We introduce a \emph{bi-semantic} training pipeline in which SMILES are not a separate semantic modality but are woven into natural-language documents at the positions of chemical entity mentions. The standard MLM objective then operates on these mixed token sequences, forcing the model to predict masked SMILES tokens from the textual context and vice versa. We subsequently use a contrastive learning objective to refine the embeddings learned during the first training phase by creating artificial SMILES-annotated text pairs through a similarity-based approach on chemical content. This training pipeline enables a single model to produce consistent representations across heterogeneous chemical and scientific domains.
  
  \textbf{(iv) Multi-purpose benchmarking.}
  To assess the flexibility of our model across the two semantics, we construct a unified benchmarking pipeline that covers 48 diverse evaluation datasets, comprising 26 classification and 22 regression tasks. These datasets span a broad range of chemistry-relevant problem settings, including molecular property prediction, materials science, and chemistry-related scientific language understanding drawn from established community benchmarks validated across literature (see Section \ref{si:table_benchmarks}).

\section{Methods}
In this section, we introduce the two-stage training pipeline that sequentially integrates masked language modeling (MLM) with a contrastive learning objective. To fuel this bi-semantic training, we processed our entire collection of 14.4 million source documents through our automated SMILES-injection pipeline. This massive source corpus comprises 6.6 million scientific texts (5.9 million abstracts and 0.9 million full-text articles) from ChemPile~\cite{mirza_chempile_2025}, as well as 7.8 million high-quality educational documents from FineWeb-EDU~\cite{penedo_fineweb_2024}. 

\subsection{Data filtering \& SMILES annotation pipeline}
\label{sec:data_annotation}
\subsubsection{Data filtering}
We collected datasets from the ChemPile corpus; chemistry papers from ChemPile-Papers (11.45B tokens), a curated, open-access collection of chemistry-related scientific articles, as well as educational texts from ChemPile-Education (75M tokens)~\cite{mirza_chempile_2025}. Additionally, our main source of educational texts is FineWeb-Edu, a 1.3T tokens subset of the FineWeb web crawl filtered using an educational content classifier trained on Llama3-70B-Instruct annotations~\cite{penedo_fineweb_2024}. As we only required chemistry-related content, we filtered the texts from FineWeb-Edu using a custom scoring pipeline: each text is first standardized (lowercase, removal of special characters) and then split into a list of words, and secondly, each text's relevance to chemistry is scored using a word frequency-based approach, using a simple word classifier algorithm. This classification allows us to compute a text chemistry score (TCS) for each text, a metric that measures the text's chemical relevance~\cite{bran_mist_2026}.
  The TCS formulas can be found in Equation~(\ref{eq:tcs}). The classification is based on the word frequencies found in two manually labeled text corpora: chemistry texts (positive) and non-chemistry texts (negative). The word $k$ frequencies in chemistry texts and non-chemistry texts are written $f^c_k$ and $f^n_k$ respectively:
\begin{equation}
\begin{split}
  TCS(\text{text}) &:= \frac{1}{N_{\text{words}}}\sum_{\substack{k=\text{word}\\\text{in text}}}w_k \\
  \text{with}\quad w_k &= 
  \begin{cases}
    f^c_k/f^n_k, & \text{if}\:\:f^c_k/f^n_k>1 \\
    0, & \text{otherwise}
  \end{cases}
\end{split}
\label{eq:tcs}
\end{equation}

  We filtered FineWeb-Edu by retaining texts with a TCS above 1, resulting in a dataset of 10.41B tokens after deduplication.
  
\subsubsection{SMILES annotation}
Each text is then processed through a four-stage SMILES injection pipeline to annotate chemical entities with their canonical SMILES representations. First, Chemical Data Extractor~2 (CDE2)~\cite{swain_chemdataextractor_2016, mavracic_chemdataextractor_2021} is used to perform chemistry-aware 
named entity recognition, detecting all chemical entities in the texts. Second, identified chemical spans are passed into OPSIN~\cite{lowe_chemical_2011} for deterministic IUPAC-to-SMILES conversion or standard name conversions. Third, each flagged chemical span that OPSIN was unable to resolve is passed to a PubChem API call via PubChemPy as a fallback lookup to obtain the associated SMILES representations~\cite{kim_pubchem_2025}. Then, all 
recovered SMILES are canonicalized and validated through RDKit~\cite{landrum2025rdkit}; structures that fail valence checking are discarded. Finally, the valid SMILES representations are inserted immediately after the original chemical entity in the source text, yielding our final annotated dataset.

\begin{figure*}[ht]
  \centering
  \includegraphics[width=\textwidth]{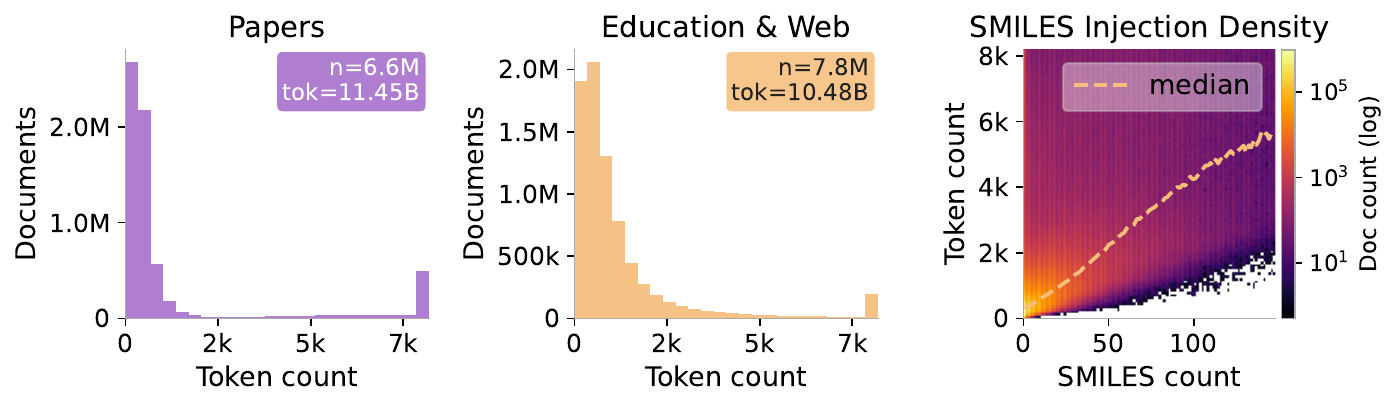}
  \caption{Distribution of sample lengths and SMILES content in the CheMatE MLM training corpus. The left and middle panels show token-count histograms for the Papers subset (6.6M documents, 11.45B tokens) and the Education/Web subset (7.8M documents, 10.48B tokens), respectively, after truncating the texts at 8,192 tokens. The right panel shows a log-scaled 2D histogram of SMILES count versus token count across the combined corpus. The dashed line represents the median token count within each SMILES-count bin.}
  \label{fig:data_distribution}
\end{figure*}

\subsection{Bi-semantic training strategies}
\subsubsection{MLM strategy}
\label{sec:implementation_details_mlm}

Each text sample consists of natural language containing SMILES strings injected using our annotation pipeline. The number of SMILES per sample is proportional to sequence length, as shown in Figure~\ref{fig:data_distribution}: the longest samples contain more than 200 SMILES within the full 8,192-token context window, providing rich co-occurrence signals among molecular species that appear in similar chemical contexts.

To infuse bi-semantic knowledge, we started by conducting a continuous pre-training of ModernBERT base~\cite{warner_smarter_2024} under a standard Masked Language Modeling (MLM) (see Equation~\ref{MLM}) objective with a 15\% masking rate, using Flash Attention 2~\cite{dao_flashattention-2_2023}~and BF16 precision across 16 GH200 GPUs (4 nodes $\times$ 4 GPUs).
 \begin{equation}
      \mathcal{L}_{\mathrm{MLM}}(\theta)
      \;=\;
      -\frac{1}{|\mathcal{M}|}
      \sum_{i\in\mathcal{M}}
      \log P_\theta\bigl(x_i \,\bigm|\, \widetilde{X}\bigr)\,
      \label{MLM}
 \end{equation}
where $\mathcal{L}_{\mathrm{MLM}}(\theta)$ is the masked language modeling loss for model weights $\theta$, $\mathcal{M}$ denotes the set of masked token positions, and $P_\theta(x_i \mid \widetilde{X})$ is the predicted probability of predicting the original token $x_i$ given the corrupted bidirectional context $\widetilde{X}$~\cite{devlin_bert_2019}.

A key challenge arises from the highly variable sequence lengths in our corpus under a multi-GPU training setup; the computational load is not evenly distributed across GPUs. Standard transformer training typically pads sequences within each batch, which implicitly homogenizes per-device compute costs. In the case of ModernBERT training, the sequences are unpadded and concatenated to optimize the training efficiency with jagged attention masks~\cite{warner_smarter_2024}. While effective in settings with homogeneous sequence lengths or limited device parallelism (e.g., single-GPU training), this approach can lead to substantial load imbalance in large-scale multi-GPU setups when sequence lengths vary significantly. In these cases, overall training efficiency decreases because some GPUs remain idle while waiting for the device with the most computationally expensive batch to complete. To address this, we created a custom batch sampler: \texttt{BalancedTokenBatchSampler}~(Algorithm~\ref{alg:smart_batching}). The objective of this sampler is to construct batches with balanced per-device computational costs, enabling efficient workload distribution across GPUs. The core principles behind the batch construction strategy is that, in ModernBERT training, the computational cost of a batch is approximately proportional to $\sum_i L_i^2$ due to the use of unpadding and jagged attention masks, whereas standard transformer training with padding scales approximately as $\sum_i L_{\max}^2$, where $L_{\max}$ denotes the length of the longest sequence in the batch. Consequently, our sampler aims to construct per-device batches such that the values of $\sum_i L_i^2$ remain balanced across GPUs at each training step.

The details of our custom batch sampler (\texttt{BalancedTokenBatchSampler}) are provided in Algorithm~\ref{alg:smart_batching}. Samples are first sorted by token length, split into $B$ bins, shuffled within each bin, and then concatenated back together. Batch stochasticity is preserved as long as $N\_samples/B$ remains sufficiently large. If $B$ becomes too large, the bin assignment becomes increasingly deterministic; however, this is not problematic in our setting, given the scale of the dataset (millions of samples) and the relatively small value of $B$. After this sample shuffling, per-device batches are formed greedily under three simultaneous constraints: a raw-token budget $T_{\max}$, a padded-token memory cap $T_{\max}^{\mathrm{pad}}$, and a quadratic cost ceiling $C_{\max}$. As a result, the sampler naturally produces variable-length per-device batches containing different numbers of sequences depending on their token lengths and estimated computational cost. The resulting per-device batches are sorted by quadratic cost, grouped into blocks of size $W$ (corresponding to the number of GPUs), and the blocks are shuffled to mitigate ordering bias. This results in a batch construction strategy that assigns per-device batches with comparable computational loads to each training step, ensuring balanced workload distribution across GPUs.

\begin{algorithm}[H]
  \caption{Smart Batching BalancedTokenBatchSampler}
  \label{alg:smart_batching}
  \begin{algorithmic}
    \STATE {\bfseries Input:} token counts $\{l_i\}_{i=1}^{N}$; $B$ bins;
    \STATE \quad $T_{\max}$, $T_{\max}^{\mathrm{pad}}$, $C_{\max}$; world size $W$, rank $r$
    \STATE {\bfseries Output:} ordered batch list for rank $r$
    \STATE Sort indices by $l_i$; split into $B$ equal bins; shuffle within each bin; concatenate to obtain $\mathcal{I}$
    \STATE $\mathcal{B} \leftarrow [\ ]$,\quad $b \leftarrow [\ ]$
    \FOR{each index $i \in \mathcal{I}$}
      \IF{$\sum_{j\in b} l_j + l_i > T_{\max}$ {\bfseries or} $\max (\max_{j \in b} l_j,\, l_i)\cdot(|b|{+}1) > T_{\max}^{\mathrm{pad}}$ {\bfseries or} $\sum_{j \in b} l_j^2 + l_i^2 > C_{\max}$}
        \STATE $\mathcal{B} \leftarrow \mathcal{B} \mathbin{\|} [b]$; \quad $b \leftarrow [\ ]$
      \ENDIF
      \STATE $b \leftarrow b \mathbin{\|} [i]$
    \ENDFOR
    \STATE Sort $\mathcal{B}$ by cost; group into $\lfloor|\mathcal{B}|/W\rfloor$ groups of size $W$
    \STATE Shuffle groups; assign group$[\,\cdot\,][r]$ to rank $r$
    \STATE \textbf{return} rank $r$'s batch list
  \end{algorithmic}
\end{algorithm}

\subsubsection{Contrastive training strategy}
\label{sec:implementation_details_cl}

To improve the bi-semantic understanding in our model and further refine the quality of the embeddings, we algorithmically constructed an artificial contrastive dataset directly from our annotated corpora by leveraging the chemical content naturally embedded in each document. The core intuition is that two text passages discussing chemically similar molecules should be close in embedding space, while passages whose molecular content is chemically unrelated should be pushed apart. Rather than relying on document-level labels, we compute this signal at the SMILES level. At each iteration, we draw $K$ documents from our collection of SMILES annotated texts $\mathcal{D}_s$ and chunk them into sentence-level segments to form a candidate pool $\mathcal{C}$. We proceed by sampling a single canonical SMILES $s^{*}$ present in a random text from $\mathcal{C}$ and use it as the anchor. For every segment $c \in \mathcal{C}$, each inline SMILES it contains is individually compared to $s^{*}$ via Tanimoto similarity over Morgan fingerprints (radius 2, 2048 bits), and the scores are aggregated (e.g.\ max) into a single segment-level score $c.\mathit{score}$~\cite{rogers_extended-connectivity_2010,landrum2025rdkit}. The top-$P$ segments whose total score exceeds $\tau^{+}$ become the positive set $\mathcal{C}^{+}$, and the bottom-$Q$ segments whose score are below $\tau^{-}$ become the negative set $\mathcal{C}^{-}$. Together with the anchor they form a triple $(s^{*}, \mathcal{C}^{+}, \mathcal{C}^{-})$, and the procedure is repeated until the contrastive dataset $\mathcal{P}$ contains $N$ such triples as illustrated on Algorithm ~\ref{alg:contrastive_pairs}. 

\begin{algorithm}[H]
  \footnotesize
  \caption{SMILES-Text Synthetic Pair generator}
  \label{alg:contrastive_pairs}
  \begin{algorithmic}
    \STATE {\bfseries Input:} corpus $\mathcal{D}_s$; size $N$; docs/round $K$;
    \STATE \quad positives $P$, negatives $Q$; thresholds $\tau^{+},\tau^{-}$
    \STATE {\bfseries Output:} dataset $\mathcal{P}=\{(s^{*},\mathcal{C}^{+},\mathcal{C}^{-})\}$
    \STATE $\mathcal{P} \leftarrow [\ ]$
    \WHILE{$|\mathcal{P}| < N$}
      \STATE draw $K$ docs from $\mathcal{D}_s$; chunk to segments $\to \mathcal{C}$
      \STATE sample $c_0 \sim \mathcal{C}$;
      \STATE \quad anchor $s^{*} \sim \mathrm{SMILES}(c_0)$
      \STATE $f^{*} \leftarrow \mathrm{Morgan}(s^{*}, r{=}2, b{=}2048)$
      \FOR{$c \in \mathcal{C}$}
        \STATE $c.\mathrm{score} \leftarrow $
        \STATE \quad $\max_{s \in \mathrm{SMILES}(c)} \mathrm{Tanimoto}(f^{*}, \mathrm{Morgan}(s))$
      \ENDFOR
      \STATE $\mathcal{C}^{+} \leftarrow \mathrm{top}_P\{c : c.\mathrm{score} > \tau^{+}\}$
      \STATE $\mathcal{C}^{-} \leftarrow \mathrm{bottom}_Q\{c : c.\mathrm{score} < \tau^{-}\}$
      \IF{$\mathcal{C}^{+} \neq \emptyset$ {\bfseries and} $\mathcal{C}^{-} \neq \emptyset$}
        \STATE $\mathcal{P} \leftarrow \mathcal{P} \mathbin{\|} [(s^{*},\mathcal{C}^{+},\mathcal{C}^{-})]$
      \ENDIF
    \ENDWHILE
    \STATE \textbf{return} $\mathcal{P}$
  \end{algorithmic}
\end{algorithm}

The model is then fine-tuned on these triples or pairs using the Multiple Negative Ranking Loss (see Equation \ref{mnrl_eq}) contrastive objective with Matryoshka loss~\cite{kusupati_matryoshka_2024}, simultaneously optimizing embedding alignment across sub-dimensions (768, 512, 256, 128, 64) to produce representations that remain informative even when truncated. In practice for CheMatE, we train for a single epoch on a 20k subset of anchor-positive pairs, which we found sufficient to achieve meaningful embedding refinement without overfitting.

The Multiple Negatives Ranking Loss (MNRL) at dimension $d_k$ is a cross-entropy objective, where corresponding texts are pulled together, and non-matching ones are pushed apart~\cite{henderson_efficient_2017}:
\begin{equation}
\label{mnrl_eq}
  \begin{split}
    \mathcal{L}_{\mathrm{MNRL}}^{(k)} = -\frac{1}{N}\sum_{i=1}^{N}\Biggl[ S_{x_i,y_i}^{(k)} - \log \sum_{j=1}^{N}e^{S_{x_i,y_j}^{(k)}} \Biggr] 
  \end{split}
\end{equation}

where $S_{x_i,y_j}^{(k)}$ is the similarity score (cosine similarity) computed between the text embeddings $x_i$ and $y_i$ at dimension $d_k$ (truncated embeddings for the Matryoshka loss). This objective encourages each anchor embedding $x_i$ to be more similar to its corresponding positive $y_i$ than to negatives $y_j$ (when $i\ne j$). The final Matryoshka loss is the weighted sum across all dimensions~\cite{kusupati_matryoshka_2024}:
\begin{equation}
    \mathcal{L}_{\mathrm{Matryoshka}} 
    = \sum_{k=1}^{K} w_k \; \mathcal{L}_{\mathrm{MNRL}}^{(k)} 
\label{matryosh_eq}
\end{equation}

\subsubsection{CheMatE training details}
\label{sub:training_details}

Our model is built on the ModernBERT-base architecture~\cite{warner_smarter_2024}, a long-context encoder-only model pre-trained with a masked language modeling objective. We perform continued MLM pre-training using \texttt{answerdotai/ModernBERT-base} native tokenizer, on our chemically-annotated corpus for 3 epochs at $\mathrm{LR}{=}5\times10^{-4}$ with a linear scheduler and 1{,}000 warmup steps, in \texttt{bf16} mixed precision. We use AdamW with $\beta_1{=}0.9$, $\beta_2{=}0.999$, $\epsilon{=}10^{-8}$, weight decay $0.01$, and gradient clipping at norm $1.0$. The MLM training is run on 16 GPUs with DDP using dynamic batch sizes. We use the final epoch-3 checkpoint as the initialization for contrastive fine-tuning. The resulting MLM checkpoint is then fine-tuned through Matryoshka contrastive learning with a Multiple Negatives Ranking Loss (MNRL) objective on pairs generated by our in-house data generation pipeline, using in-batch negatives and mean token pooling to obtain sequence embeddings during training. The synthetic pair dataset generated was filtered to keep only anchor SMILES of length $\geq 64$ characters (19{,}638 pairs out of an initial 100{,}000), and the model was trained for 3 epochs at $\mathrm{LR}{=}2\times10^{-5}$ with 500 linear warmup steps followed by cosine decay. The contrastive training is run on 4 GPUs with DDP using a per-rank batch size of 16 and gradient accumulation of 2, yielding an effective batch size of 128 (around 15 in-batch negatives per anchor per rank). The released CheMatE checkpoint corresponds to the end of the first contrastive epoch. For evaluation, we employ mean-token pooling to match our training methodology.

\section{Results}
We evaluate CheMatE against a total of 11 baselines on 48 benchmark datasets (26 classification, 22 regression) drawn from a diverse collection spanning molecular property prediction (see Section~\ref{si:table_benchmarks}), materials science, and chemistry-related NLP tasks~\cite{gurulingappa_development_2012, ramakrishnan_quantum_2014, bravo_extraction_2015, baker_automatic_2016, wu_moleculenet_2018, cohan_structural_2019, arxiv_cat_2019, jin_pubmedqa_2019, kotonya_explainable_2020, huang_therapeutics_2021, singh_scirepeval_2022, wognum_call_2024, herck_assessment_2025, kasmaee_chemteb_2025}. For comparison purposes, we select baselines containing both general-purpose language models (ModernBERT~\cite{warner_smarter_2024}, SciBERT~\cite{beltagy_scibert_2019}, Nomic-v1.5~\cite{nussbaum_nomic_2025}, GTE-base-v1.5~\cite{zhang_mgte_2024}, Nomic-MLM, and GTE-MLM~\cite{zhang_mgte_2024}) and chemistry-specialized encoder-based models (ChemBERTa-77M-MLM~\cite{chithrananda_chemberta_2020}, MoLFormer-c3-1.1B~\cite{singh_chemberta-3_2026}, BERT-SMILES~\cite{jouary_bridging_2025}, and ChEmbed-full~\cite{kasmaee_chembed_2025}. Additionally, we include a standard Morgan fingerprint (radius 2, 2048 bits) baseline~\cite{rogers_extended-connectivity_2010} for SMILES-only tasks to compare our results against established cheminformatics representations. To assess the embedding quality of the text provided, we adopt a frozen-embedding evaluation protocol. A linear probe is initialized on top of the fixed embedding, ensuring that the transformer backbones act purely as feature extractors without any task-specific fine-tuning.
For classification tasks, we employ a multinomial logistic regression classifier with balanced class weights and report balanced accuracy as our main evaluation metric. For regression tasks, we train a Ridge regressor ($\alpha=1.0$) and report $R^2$ as our primary metric.

\begin{figure*}[ht]
  \centering
  \includegraphics[width=\textwidth]{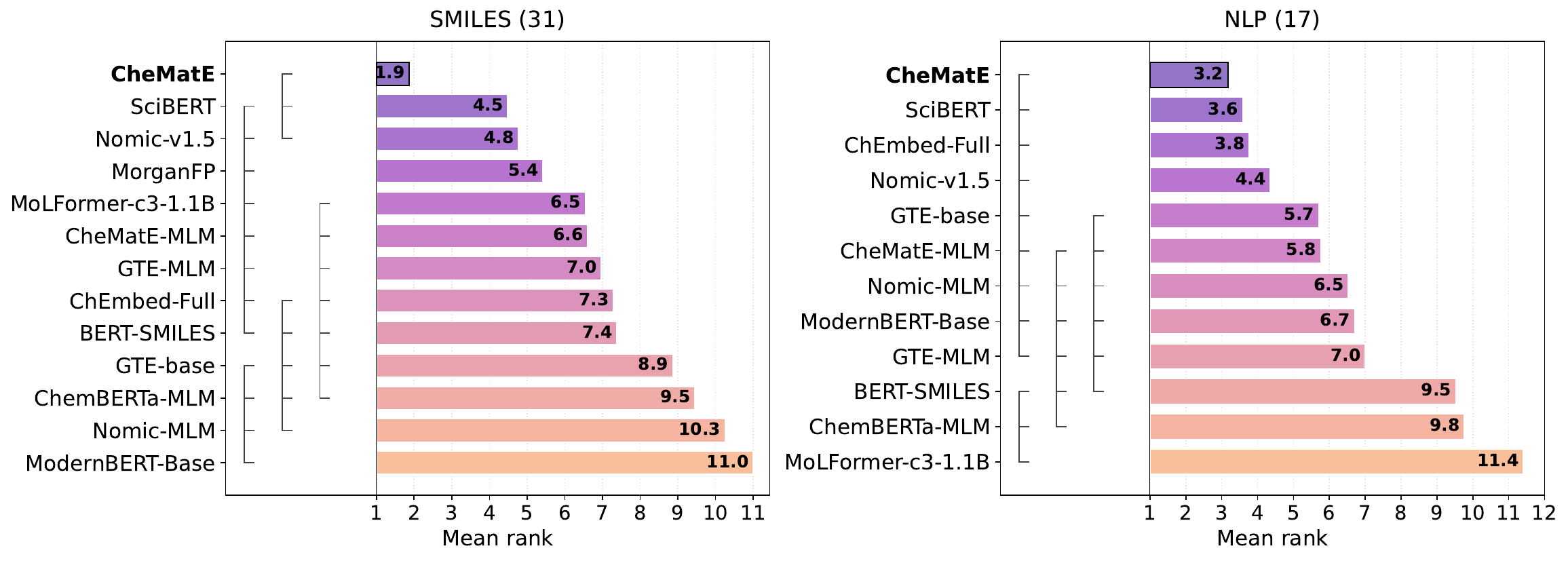}
  \caption{
    Per-modality Demšar critical-difference (CD) diagrams over 20 cross-validation folds (Friedman omnibus + post-hoc Nemenyi, $\alpha{=}0.05$)~\cite{friedman_use_1937, demsar_statistical_2006}. Each bar shows a model's mean rank on fold-averaged scores across the given datasets (lower is better, where 1 is the best). Vertical connector bars join models whose pairwise rank differences are not significant under the Nemenyi test. The metrics used for classification and regression are, respectively, balanced accuracy and $R^{2}$. Both plots show aggregated results per representation with 13 regression and 18 classification SMILES-based tasks (left panel), and 4 regression with 13 classification datasets for scientific natural language (right panel). 
}
  \label{fig:cd_split_1x2}
\end{figure*}

Figure~\ref{fig:cd_split_1x2} displays aggregated Critical Difference (CD) diagrams for both SMILES and scientific natural language modalities, pooling classification and regression tasks to maximize statistical power per panel. We use the Friedman omnibus test to reject the null hypothesis of equal model performance, and the Nemenyi post-hoc test ($\alpha{=}0.05$) to identify pairwise equivalences~\cite{friedman_use_1937, demsar_statistical_2006}. Models with no significant difference are connected by vertical brackets. Across both semantic modalities, CheMatE achieves the lowest mean rank ($1.9$ on SMILES, $3.2$ on NLP) and is the only chemistry-specialized model in the top Nemenyi equivalence group on both panels. 

This result demonstrates the strong bi-semantic capabilities of our model without the modality trade-off present in SMILES-only encoders such as MoLFormer and BERT-SMILES. In addition, our mid-stage checkpoint (CheMatE-MLM) ranks significantly lower than our final contrastively tuned model on the SMILES panel (rank $6.6$ vs $1.9$), indicating that the second sequential training step crucially refines embedding quality. Furthermore, CheMatE outranks Morgan fingerprints as well as chemistry-specialized encoders such as MoLFormer-c3 and ChemBERTa-MLM. On the NLP panel ($n{=}17$), CheMatE also achieves the lowest mean rank ($3.2$) and is not significantly worse than any dedicated scientific language model, supporting that chemical specialization does not come at the cost of natural scientific language representational quality.

However, the true advantage of CheMatE is revealed by the cross-modal asymmetry between the two panels. While some text-trained encoders manage to retain competitive performances on SMILES tasks (serving as CheMatE's closest competitors), every chemistry-specialized SMILES baseline, including MoLFormer-c3-1.1B ($11.4$), ChemBERTa-MLM ($9.8$), and BERT-SMILES ($9.5$), is located at the bottom of the NLP leaderboard. CheMatE is the only model that avoids this semantic trade-off, confirming that our joint contrastive training objective on our in-house annotated data effectively produces a single encoder capable of robust performance across both representations.

To complement this view, we employ a per-dataset frequency-of-best group analysis following the principles of~\citet{ash_practically_2025}. For each dataset in our collection, we conduct a one-way ANOVA across all models over the 20 CV folds, and apply a Tukey's honestly significant difference (HSD) post-hoc test on those datasets where the omnibus test rejects equal-means at $\alpha{=}0.05$~\cite{tukey_comparing_1949, ash_practically_2025}. When the test fails to reject the null hypothesis, we conservatively treat all models as tied, counting each as belonging to the best-performing group for that dataset.

\begin{table*}[ht]
\centering
\caption{Frequency of models in the best-performing statistical group across all evaluated datasets. The best-performing group is defined as the top-scoring model and any models statistically indistinguishable from it under a Tukey's HSD post-hoc test ($\alpha=0.05$, 20 CV folds). If the initial ANOVA omnibus test lacked significance, all models were considered tied. Results are displayed as raw counts (in-best group/N$_{\text{benchmarks}}$) per modality, along with a bi-semantic score (\%) built by weighting the two modalities equally. We denote our models with an asterisk ($^\ast$).}
\label{tab:tukey_combined}
\setlength{\tabcolsep}{4pt}
\begin{tabular}{lccr}
\toprule
& \multicolumn{2}{c}{In-best group} & \\
\cmidrule(lr){2-3}
Model & SMILES & NLP & Bi-sem. \% \\
\midrule
CheMatE$^\ast$        & 30/31 & 13/17 & \textbf{86.7} \\
SciBERT               & 23/31 & 13/17 & 75.4 \\
Nomic-v1.5            & 23/31 & 10/17 & 66.5 \\
ChEmbed-Full          & 17/31 & 10/17 & 56.8 \\
CheMatE-MLM$^\ast$    & 18/31 &  9/17 & 55.5 \\
MoLFormer-c3-1.1B     & 24/31 &  4/17 & 50.5 \\
GTE-MLM               & 20/31 &  6/17 & 49.9 \\
GTE-base-v1.5         & 14/31 &  8/17 & 46.2 \\
BERT-SMILES           & 18/31 &  4/17 & 40.8 \\
Nomic-BERT-MLM        & 10/31 &  8/17 & 39.7 \\
ModernBERT-Base       & 11/31 &  7/17 & 38.4 \\
ChemBERTa-MLM         & 10/31 &  5/17 & 30.9 \\
Morgan FP (r=2, 2048) & 21/31 & ---   & ---  \\
\bottomrule
\end{tabular}
\end{table*}

Table~\ref{tab:tukey_combined} reports the proportion of datasets where each model ranks in the top tier, which is defined as being statistically indistinguishable from the best-performing model. While the CD diagram illustrates average ordering across datasets, this analysis offers a view of the frequency at which each model is among the dataset's top performers. We design a new evaluation metric, named \textit{Bi-semantic Score \%}, to account for dataset modality imbalance, by weighting the two modalities equally. 

Across all baselines, CheMatE achieves a bi-semantic score of $86.7\%$ and belongs to the best-performing group on $87.8\%$ of datasets (i.e., the best group in $43/48$ datasets), with consistent top-tier performance across modalities: $30/31$ on SMILES and $13/17$ on NLP. This produces a single encoder that is competitive on chemistry and scientific natural language with minimal observed specialization trade-off. Comparing CheMatE to its MLM-only counterpart enables isolating the contribution of the contrastive training stage: removing the contrastive supervision reduces the bi-semantic score to $55.5\%$ ($27/48$ datasets), a $31.2\%$ performance difference. This ablation indicates that contrastive training over scientific corpora is also a key ingredient driving CheMatE's cross-modality performance.

\section{Discussion}

We presented CheMatE, a bi-semantic chemistry encoder that jointly represents molecular SMILES and scientific natural language within a unified embedding space. Our model mitigates the trade-off between chemical and natural language specialization that is observed in current encoder models by combining a large-scale SMILES injection pipeline to generate high-quality annotated texts and a two-stage sequential training strategy. CheMatE obtains a bi-semantic score of 86.7\% and has the lowest mean rank across both SMILES and natural language modalities when evaluated on 48 benchmark datasets covering molecular property and scientific NLP predictive tasks. Our results demonstrate that strong SMILES and scientific language understanding are not mutually exclusive in encoder-only models. Furthermore, our ablation study shows that the contrastive stage is critical as the MLM-only checkpoint (CheMatE-MLM) ranks significantly lower despite sharing the same backbone. 

In addition to the model itself, this work provides an automatic SMILES annotation pipeline for scientific texts and introduces a collection of 14.4 million annotated documents spanning from chemistry papers to educational web content. To train efficiently on this heterogeneous corpus with a multi-GPU setup, we developed the \texttt{BalancedTokenBatchSampler}, a cost-aware dynamic batch sampler that enforces simultaneous constraints on token budget, memory, and quadratic computational cost across distributed GPU ranks, reducing padding overhead by up to 50\% and enabling stable MLM training at 8,192-token context lengths. We further introduce an automated method for constructing contrastive training pairs directly from annotated corpora by scoring the chemical content of annotated text chunks, thereby reducing the reliance on expensive labeling methodologies.  
Despite these contributions, several limitations remain and motivate future work. First, our SMILES annotation pipeline depends on a cascade of NER (CDE2) \cite{mavracic_chemdataextractor_2021}, rule-based parsing (OPSIN) \cite{lowe_chemical_2011}, and database lookup (PubChem)\cite{kim_pubchem_2025}, which can fail on novel compounds, ambiguous names, unusual text formatting, or entities described only by partial structural fragments. Although RDKit canonicalization \cite{landrum2025rdkit} filters invalid structures, silent annotation errors (e.g., the wrong tautomer or stereoisomer) can propagate into the training corpus and are challenging to quantify at scale. Second, our contrastive stage uses a relatively modest 20k filtered anchor-positive pairs. The trade-off between dataset size, similarity threshold for building positive/negative pairs, and downstream performance has not been systematically explored. Third, while CheMatE supports an 8,192-token context window, most downstream benchmarks considered in this study consist primarily of short SMILES strings or short scientific passages, meaning that the model's long-context capabilities are not directly stressed in our evaluation. 

Nevertheless, we believe CheMatE supports the idea that merging molecular notation directly into scientific text is a fruitful route toward general-purpose chemistry representation learning for encoders and that structural and contextual chemical understanding are not mutually exclusive goals. Beyond feature extraction, the contrastive training stage further positions CheMatE as a promising backbone for retrieval applications in mixed-semantic environments. We also hypothesize that bi-semantic encoder representations of this kind could serve as a foundation for future generative chemistry models, where grounding molecular structures in rich scientific contexts through Retrieval-Augmented Generation (RAG) pipelines may prove particularly valuable.

\section*{Software and Data}
Custom code and models developed for this work are implemented in Python and are publicly available in an open-source GitHub repository at \url{https://github.com/schwallergroup/CheMatE} under the MIT license and at \url{https://huggingface.co/SchwallerGroup/CheMatE-v0}.

\section*{Impact Statement}

This work advances representation learning at the interface of chemistry and natural language, with potential applications in drug discovery, materials science, and scientific literature. There are many potential societal consequences of our work, none which we feel must be specifically highlighted here. 

\section*{Acknowledgments}
The authors gratefully acknowledge the financial support of the Swiss National Science Foundation (SNSF). This work was also supported under project ID a131 as part of the Swiss AI Initiative, through a grant from the ETH Domain and computational resources provided by the Swiss National Supercomputing Centre (CSCS) under the Alps infrastructure.

\end{multicols}

\bibliography{chemate}
\bibliographystyle{plainnat}


\newpage
\appendix
\onecolumn
\section{Appendix}

\subsection{Details on benchmarks collected}
\label{si:table_benchmarks}

\begin{table}[ht]
  \centering
  \small
  \caption{Full list of the 48 datasets in our evaluation suite, organized by source collection. Each dataset is evaluated under 20-fold cross-validation (StratifiedKFold for classification, KFold for regression).}
  \label{tab:benchmark-datasets-full}
  \renewcommand{\arraystretch}{1.05}
  \begin{tabular}{p{0.27\textwidth} p{0.10\textwidth} p{0.55\textwidth}}
    \toprule
    \textbf{Collection} & \textbf{Task} & \textbf{Datasets} \\
    \midrule
    MoleculeNet (SMILES)         & Cls.  & BACE, HIV \\
    MoleculeNet (SMILES)        & Reg.  & ESOL, FreeSolv, Lipophilicity, QM9 (gap), QM9 (zpve) \\
    \midrule
    Polaris TDC (SMILES)        & Cls.  & Ames Mutagenicity, BBB Penetration, CYP2C9 Inhibition,
                                          CYP2D6 Inhibition, CYP3A4 Inhibition, CYP3A4 Substrate,
                                          DILI, hERG Cardiotoxicity \\
    Polaris TDC (SMILES)        & Reg.  & Caco2 Permeability, LD50 (Zhu), Lipophilicity (AZ) \\
    \midrule
    Polaris Certified (SMILES)  & Reg.  & ASAP ADMET - LogD, ASAP ADMET - MDR1-MDCKII,
                                          ASAP ADMET - MLM, Biogen ADME - HLM CLint,
                                          Biogen ADME - MDR1-MDCK ER \\
    \midrule
    GPT Challenge (SMILES)      & Cls.  & Cycloadd.\ Energy, Densities (Monomer), Synthesability \\
    GPT Challenge (SMILES)      & Reg.  & Free E.\ cyclo, MP tryg., Monomer Densities,
                                          Monomer Ecoh, Monomer Tg \\
    \midrule
    GPT Challenge (NLP)         & Cls.  & CO\textsubscript{2} ads.\ of Bio-Adsorbents,
                                          Carbondioxide Adsorption, Gasification Biomass,
                                          Thermal Desalination 2 \\
    GPT Challenge (NLP)         & Reg.  & Carbondioxide Adsorption \\
    \midrule
    NLP Scientific              & Cls.  & ADE Corpus, ArXiv Categories, GAD Gene-Disease,
                                          Hallmarks of Cancer, Health Fact,
                                          SciCite Citation Intent, SciRepEval Field of Study \\
    \midrule
    NLP Biomedical              & Cls.  & PubMedQA \\
    \midrule
    NLP Materials               & Reg.  & AFLOW Band Gap (eV), TextEdge Band Gap (eV),
                                          TextEdge Volume (log A\textsuperscript{3}) \\
    \midrule
    ChemTEB                     & Cls.  & WikipediaChemFields \\
    \bottomrule
  \end{tabular}
\end{table}

\subsection{Tukey's HSD Test Individual Dataset Performances}
\begin{figure*}[htbp]
  \centering
  \includegraphics[width=\textwidth, height=0.85\textheight, keepaspectratio]{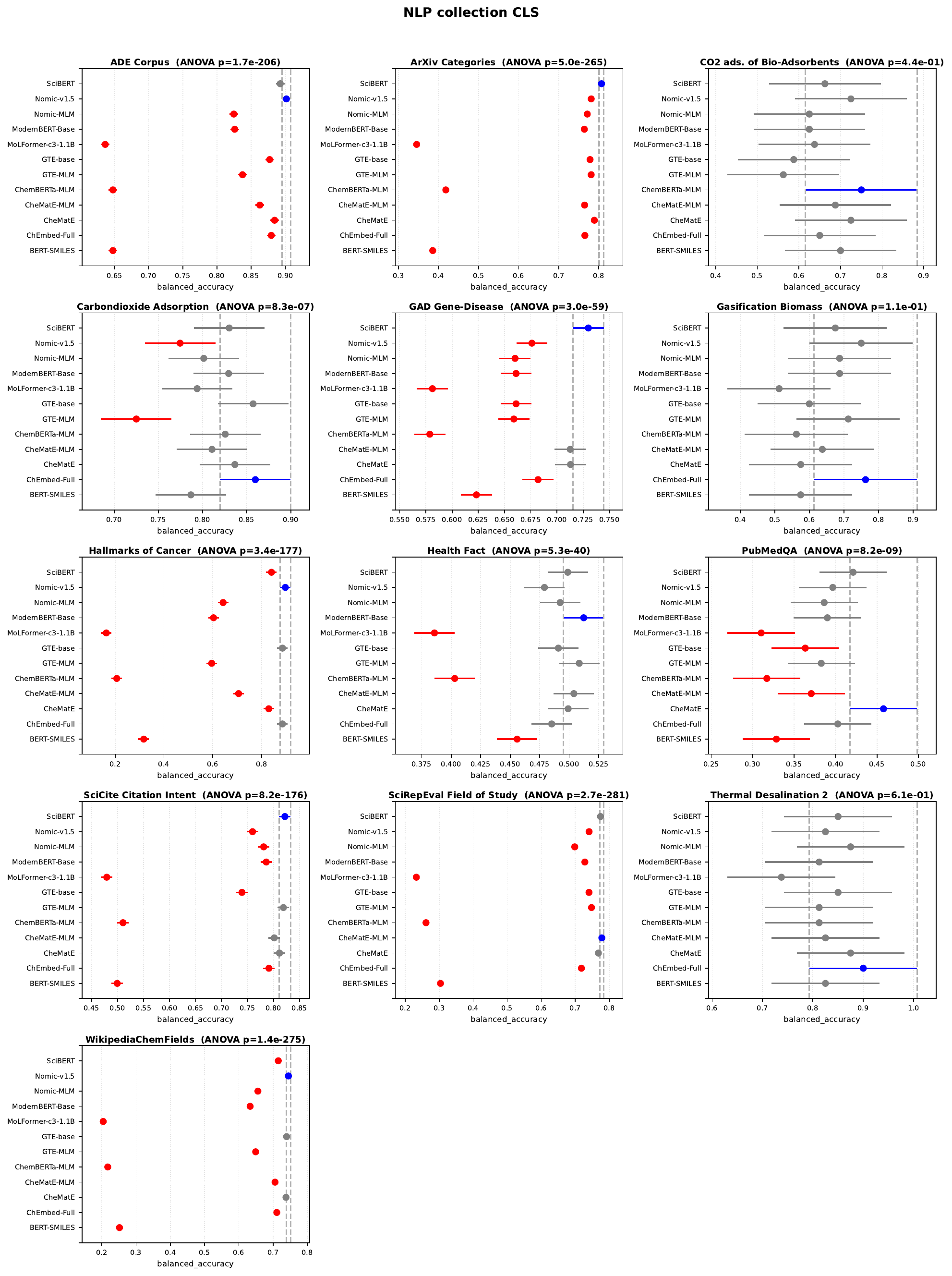}
  \vspace{0.1in}
  \begin{minipage}{\textwidth}
    \caption{Tukey HSD simultaneous confidence-interval plots for the 13 NLP classification datasets, using balanced accuracy over 20 CV folds. Each panel shows the average performance for each model and its Tukey HSD interval relative to the best mean model for that dataset~\cite{ash_practically_2025}. Blue marks the best mean, gray indicates models not statistically significantly different from the best, and red indicates models significantly worse than the best. The sub-plot titles report the one-way ANOVA p-value used to justify the post-hoc comparison.}
    \label{fig:walters_cls_nlp}
  \end{minipage}
\end{figure*}

\begin{figure*}[ht]
  \centering
  \includegraphics[width=\textwidth]{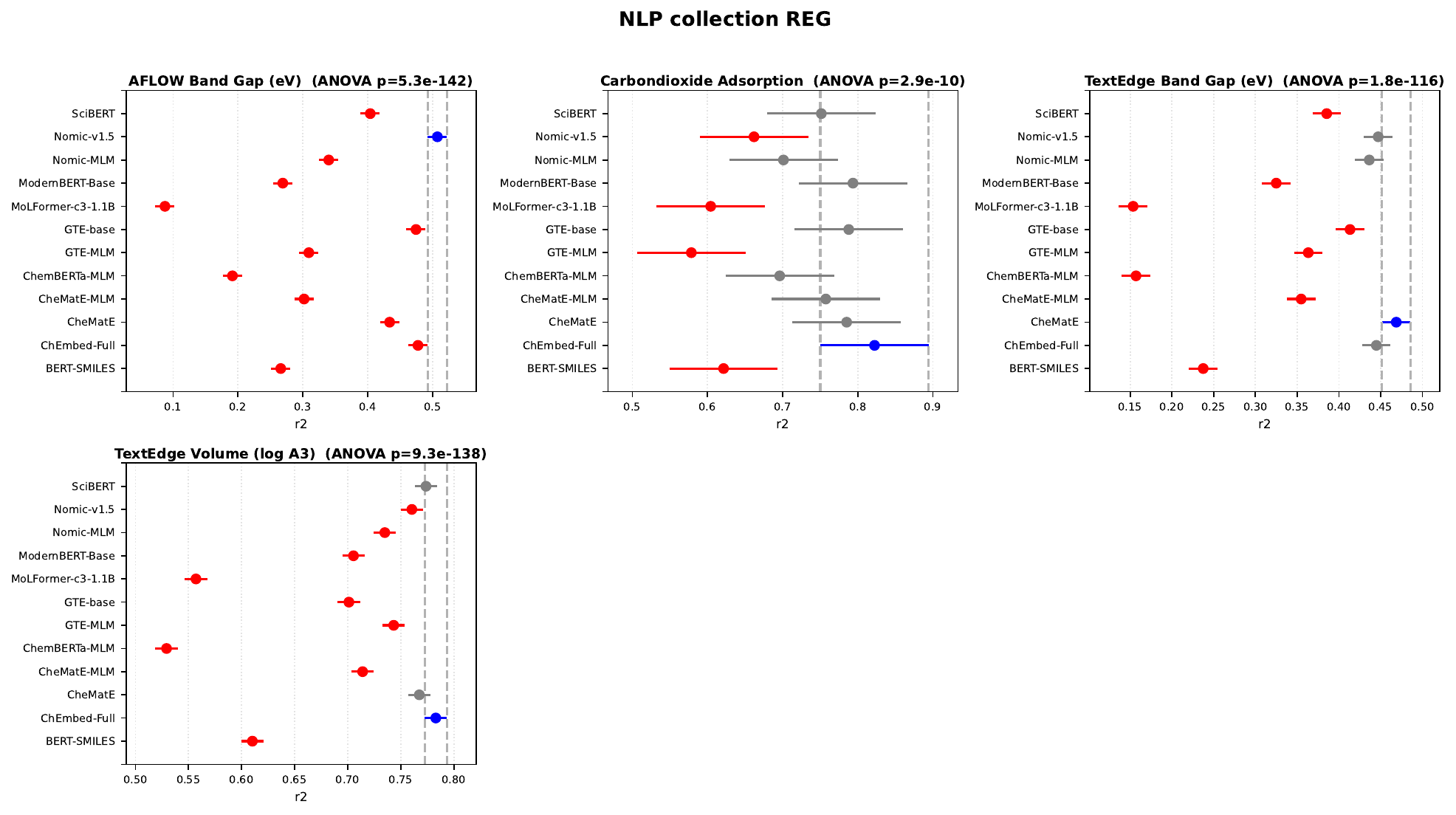}
  \caption{
    Tukey HSD simultaneous confidence-interval plots for the 4 NLP regression datasets, using balanced accuracy over 20 CV folds. Each panel shows the average performance for each model and its Tukey HSD interval relative to the best mean model for that dataset~\cite{ash_practically_2025}. Blue marks the best mean, gray indicates models not statistically significantly different from the best, and red indicates models significantly worse than the best. The sub-plot titles report the one-way ANOVA p-value used to justify the post-hoc comparison.
}
  \label{fig:walters_reg_nlp}
\end{figure*}

\begin{figure*}[htbp]
  \centering
  \includegraphics[width=\textwidth, height=0.85\textheight, keepaspectratio]{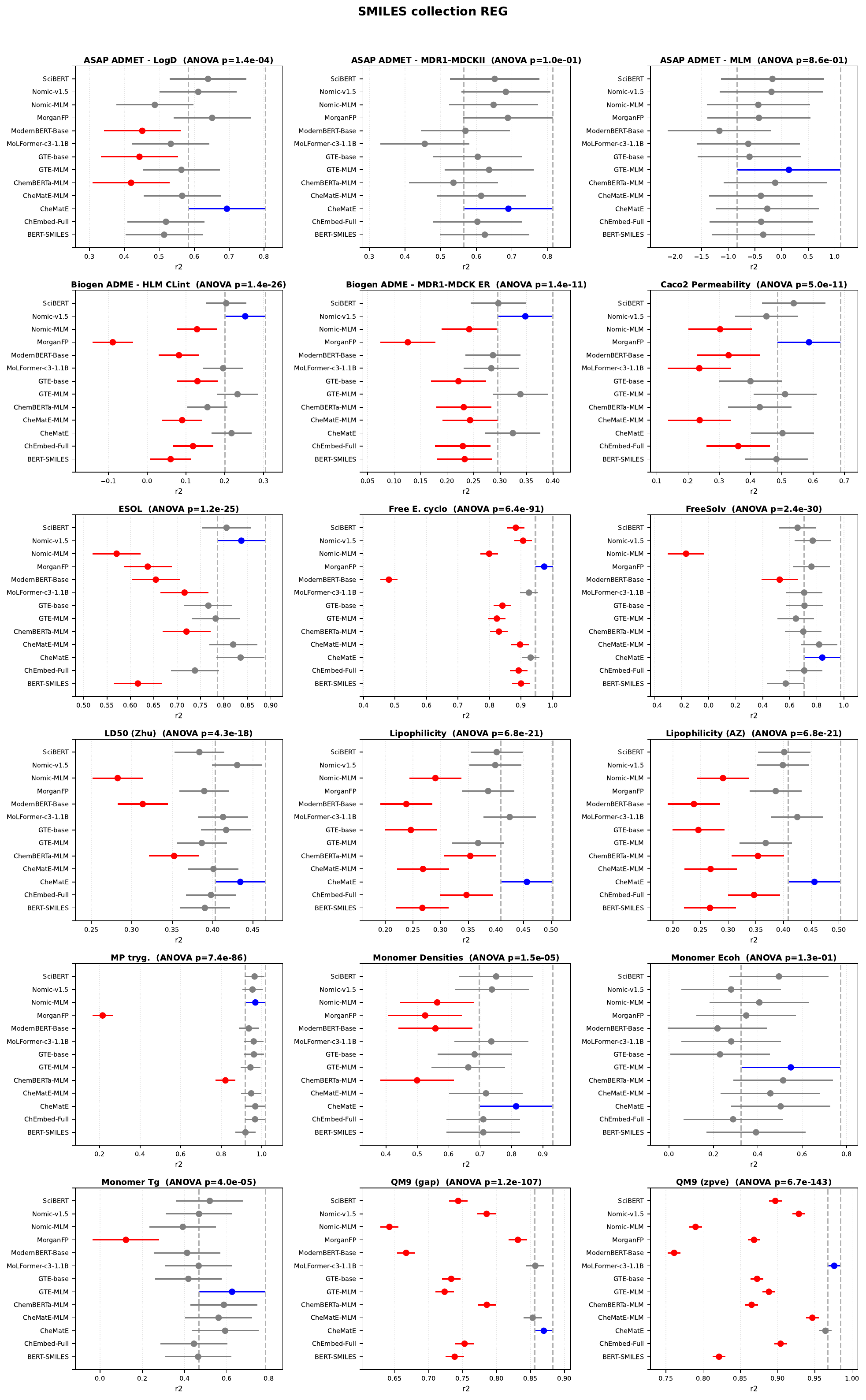}
  
  \vspace{0.1in}
  \begin{minipage}{\textwidth}
    \caption{Tukey HSD simultaneous confidence-interval plots for the 18 SMILES regression datasets, using balanced accuracy over 20 CV folds. Each panel shows the average performance for each model and its Tukey HSD interval relative to the best mean model for that dataset~\cite{ash_practically_2025}. Blue marks the best mean, gray indicates models not statistically significantly different from the best, and red indicates models significantly worse than the best. The sub-plot titles report the one-way ANOVA p-value used to justify the post-hoc comparison.}
    \label{fig:walters_reg_smiles}
  \end{minipage}
\end{figure*}

\begin{figure*}[htbp]
  \centering
  \includegraphics[width=\textwidth, height=0.85\textheight, keepaspectratio]{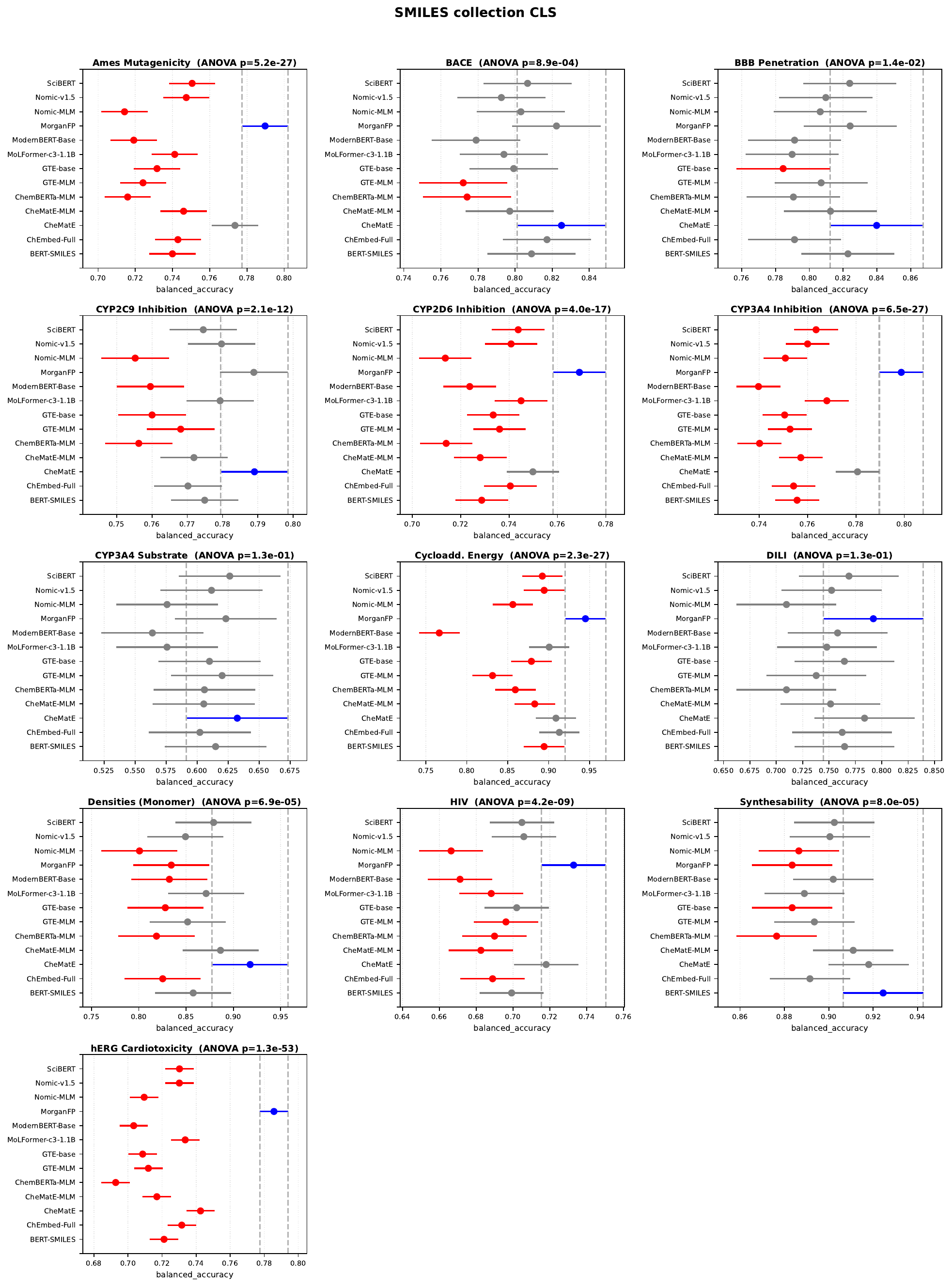}
  \vspace{0.1in}
  \begin{minipage}{\textwidth}
    \caption{Tukey HSD simultaneous confidence-interval plots for the 13 SMILES classification datasets, using balanced accuracy over 20 CV folds. Each panel shows the average performance for each model and its Tukey HSD interval relative to the best mean model for that dataset~\cite{ash_practically_2025}. Blue marks the best mean, gray indicates models not statistically significantly different from the best, and red indicates models significantly worse than the best. The sub-plot titles report the one-way ANOVA p-value used to justify the post-hoc comparison.}
    \label{fig:walters_cls_smiles}
  \end{minipage}
\end{figure*}

The supplementary Tukey's HSD test plots can be found in Figures~\ref{fig:walters_cls_nlp}, \ref{fig:walters_reg_nlp}, \ref{fig:walters_reg_smiles}, and \ref{fig:walters_cls_smiles}.

\subsection{Ranking results for each semantic modalities and each objective task}
\begin{figure*}[ht]
  \centering
  \includegraphics[width=\textwidth]{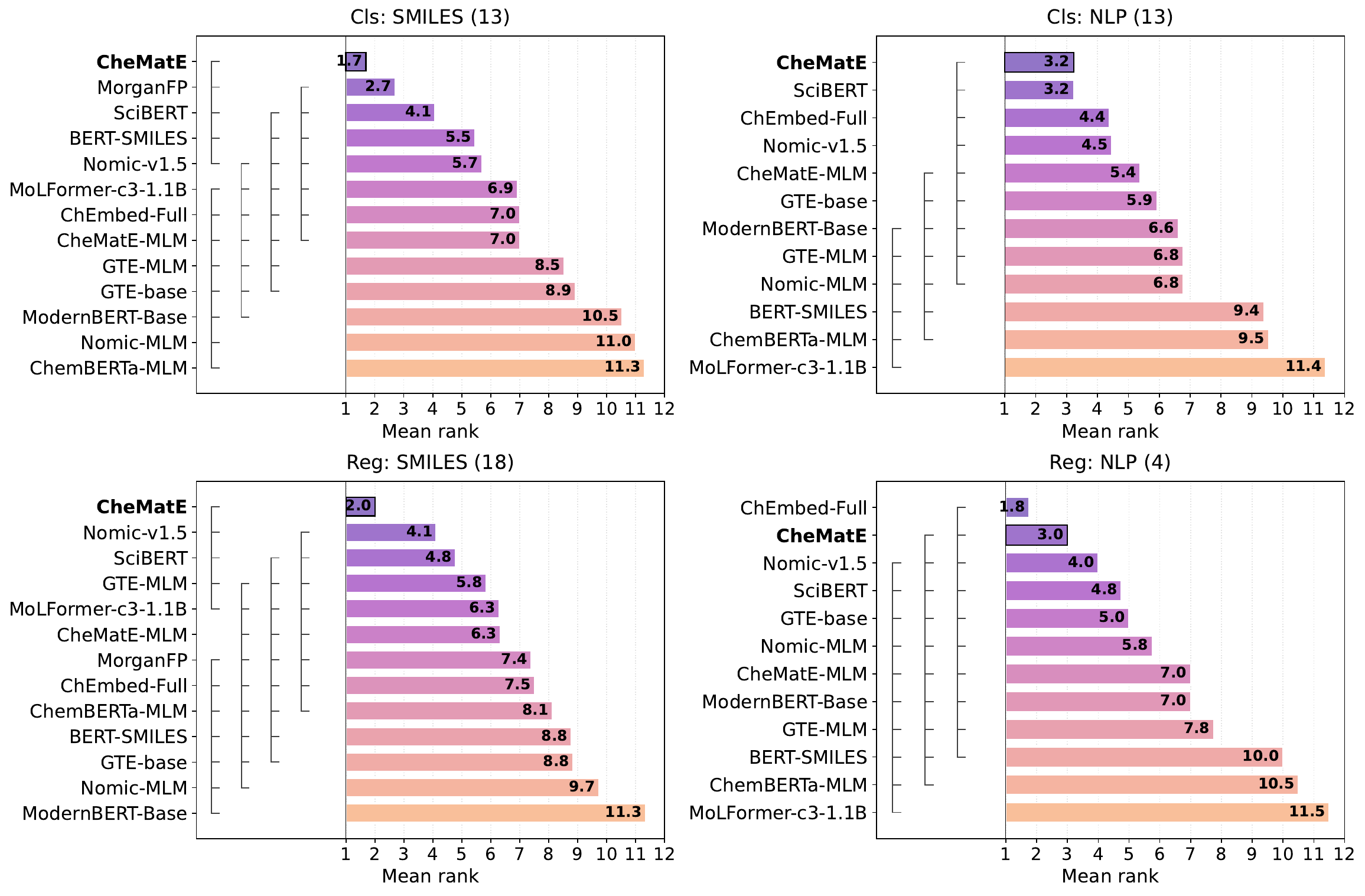}
  \caption{
    Per-modality Demšar critical-difference (CD) diagrams over 20 Cross-Validation folds (Friedman omnibus + post-hoc Nemenyi, $\alpha{=}0.05$). Each bar shows a model's mean rank on fold-averaged scores across the given datasets (lower is better, where 1 is the best). Vertical connector bars join models whose pairwise rank differences are not significant under the Nemenyi test. The metrics used for classification and regression are, respectively, balanced accuracy and $R^{2}$. Panel sizes; \textit{Cls: SMILES} 13 datasets, \textit{Cls: NLP} 13 datasets, \textit{Reg: SMILES} 18 datasets, \textit{Reg:  NLP} 4: datasets ($N{=}48$ in total). 
}
  \label{fig:cd_split_2x2}
\end{figure*}

The detailed ranking results can be found in Figure~\ref{fig:cd_split_2x2}.

\subsection{Baseline model loading}
We compare CheMatE against 11 baselines from BERT-style chemistry encoders, BERT-style general-purpose encoders, modern open sentence-embedding models, and a non-transformer molecular featuriser. All transformer baselines are loaded through HuggingFace's \texttt{AutoModel} functionality, retain their released weights, and have \emph{no} additional layers added at evaluation time. For each baseline, we apply a single pooling operation to the final hidden states, followed by L2 normalization. Pooling is selected by a per-model rule. When the released model card or repository explicitly mentions a pooling strategy (e.g., contrastive-trained sentence encoders), we use the documented choice. Otherwise, we follow the convention for the model's architecture, applying CLS-token pooling to masked-language-model checkpoints and mean-token pooling to sentence-embedding models. The models used as baselines for MLM are the following: \textbf{ModernBERT-base}~\cite{warner_smarter_2024} (\texttt{answerdotai/ModernBERT-base}) is the long-context encoder we build upon. \textbf{ChemBERTa-MLM} (\texttt{DeepChem/ChemBERTa-77M-MLM}) is 77M-parameter RoBERTa-style chemistry encoders pre-trained respectively on SMILES MLM ~\cite{chithrananda_chemberta_2020}. \textbf{BERT-SMILES} (\texttt{unikei/bert-base-smiles}) is a BERT-base model pre-trained from scratch on SMILES strings~\cite{jouary_bridging_2025}. \textbf{SciBERT} (\texttt{allenai/scibert\_scivocab\_uncased}) is BERT-base pre-trained on scientific text with a domain-specific vocabulary~\cite{beltagy_scibert_2019}. \textbf{MoLFormer-c3-1.1B} (\texttt{DeepChem/MoLFormer-c3-1.1B}) is a 1.1B SMILES encoder with linear attention~\cite{singh_chemberta-3_2026}. \textbf{Nomic-BERT-MLM} (\texttt{nomic-ai/nomic-bert-2048}) is the 2048-token Nomic-BERT MLM checkpoint~\cite{nussbaum_nomic_2025}. \textbf{GTE-MLM} (\texttt{Alibaba-NLP/gte-en-mlm-base}) is the base-MLM checkpoint released with the GTE family~\cite{zhang_mgte_2024}. For contrastive models, \textbf{Nomic-v1.5} (\texttt{nomic-ai/nomic-embed-text-v1.5}) is a trained through contrastive learning 137M-parameter sentence encoder~\cite{nussbaum_nomic_2025}. \textbf{GTE-base-v1.5} (\texttt{Alibaba-NLP/gte-base-en-v1.5}) is the base sentence encoder of the GTE family~\cite{zhang_mgte_2024}. \textbf{ChEmbed-full} (\texttt{BASF-AI/ChEmbed-full}) is a chemistry sentence-embedding model~\cite{kasmaee_chembed_2025}. \textbf{CheMatE-MLM} is our own MLM-only checkpoint (the contrastive ablation, with the same architecture as CheMatE but without contrastive fine-tuning.

\end{document}